\documentclass{article}
\usepackage{spconf,amsmath,epsfig}

\usepackage{amsfonts}
\begin{document}\sloppy

\def\x{{\mathbf x}}
\def\L{{\cal L}}

\title{Everyone is a Cartoonist: Selfie Cartoonization\\
with Attentive Adversarial Networks}
\name{Xinyu Li, Wei Zhang, Tong Shen, Tao Mei}
\address{JD AI Research, Beijing, China\\
lixinyu6@jd.com,\ wzhang.cu@gmail.com,\  shentong7@jd.com,\ tmei@live.com}

\maketitle

\begin{abstract}
Selfie and cartoon are two popular artistic forms that are widely presented in our daily life. Despite the great progress in image translation/stylization, few techniques focus specifically on selfie cartoonization, since cartoon images usually contain artistic abstraction (e.g., large smoothing areas) and exaggeration (e.g., large/delicate eyebrows). In this paper, we address this problem by proposing a selfie cartoonization Generative Adversarial Network (scGAN), which mainly uses an attentive adversarial network (AAN) to emphasize specific facial regions and ignore low-level details. More specifically, we first design a cycle-like architecture to enable training with unpaired data. Then we design three losses from different aspects. A total variation loss is used to highlight important edges and contents in cartoon portraits. An attentive cycle loss is added to lay more emphasis on delicate facial areas such as eyes. In addition, a perceptual loss is included to eliminate artifacts and improve robustness of our method. Experimental results show that our method is capable of generating different cartoon styles and outperforms a number of state-of-the-art methods.
\end{abstract}
\begin{keywords}
Self cartoonization, generative adversarial network, attention mechanism, image translation
\end{keywords}
\section{Introduction}
\begin{figure}[!htbp]
\begin{center}
\includegraphics[width=80mm]{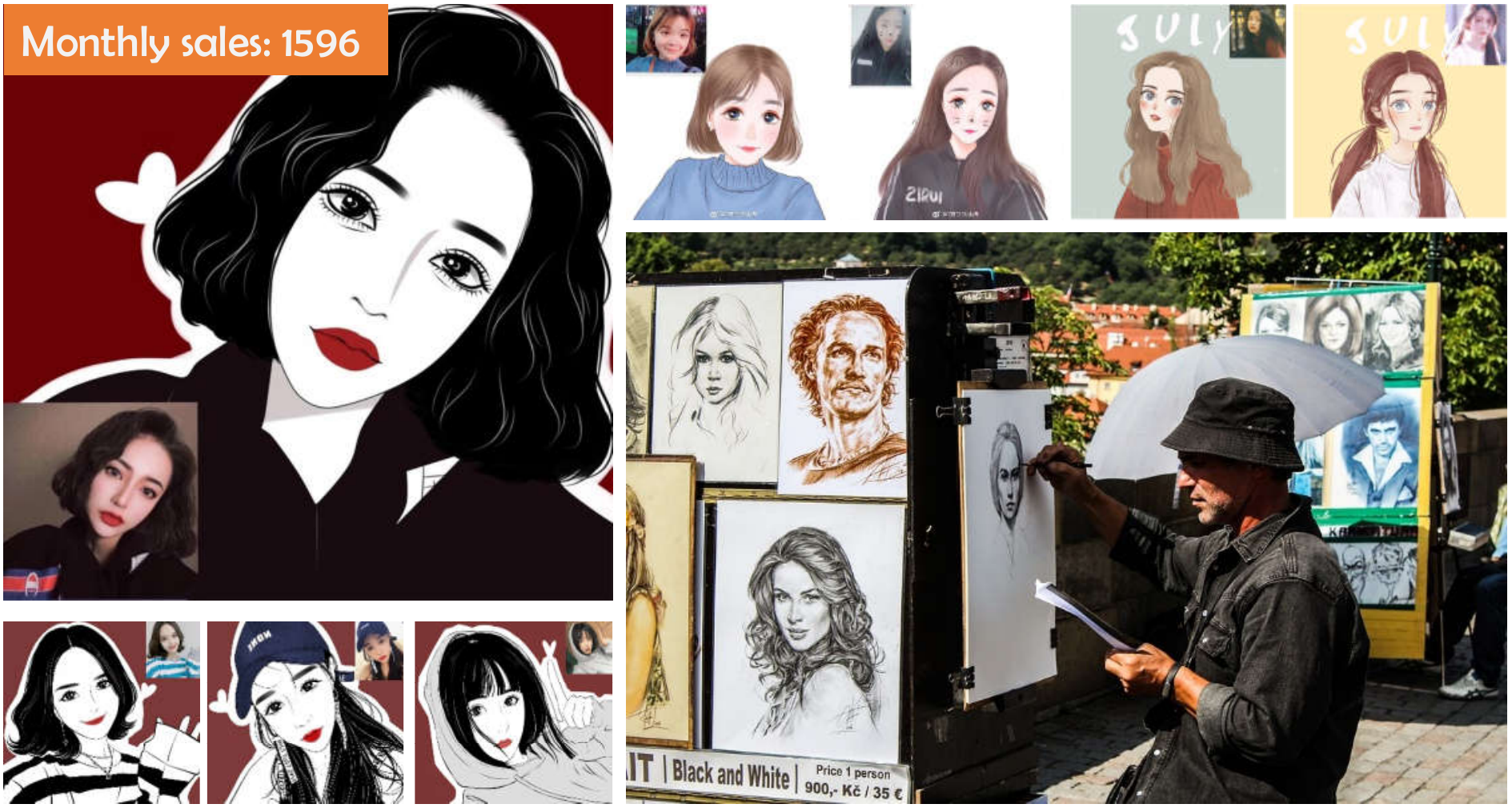}
\caption{This figure presents some cartoon portrait artworks in different styles drawn by painters. It usually takes 2 to 3 days for such a complex manual creative process.}
\end{center}
\label{fig:mot}
\vspace{-8mm}
\end{figure}

Selfie cartoonization as an artistic form is in great demand in our daily life. The most common is served as a profile in social networks, which can catch one's attention at once in such a humorous way and protect individual privacy simultaneously. In addition, the cartoon portraits are also widely used in online role-playing games, artistic poster designs and so on. However, as shown in Fig.~\ref{fig:mot}, manually drawing a cartoon portrait is very laborious and involves substantial artistic skills even with photo editing software. Thus, how to make selfie cartoonization efficient and high-quality is an important question.

Existing methods attempt to do various painting styles for cartoon portrait generation. Traditional image processing methods based on sketch extracting~\cite{canny1987computational} with little postprocessing on colors or shapes have been widely applied in smartphone software. These approaches often need to design dedicated algorithms for specific styles, and the quality of these synthetic results is far from satisfactory on a fine-grained level. The recent emergence of deep convolutional neural networks~\cite{deeplearning} has provided some attractive solutions for domain transfer. Neural Style Transfer (NST)~\cite{nst} is one of them, which is able to transfer the artistic style of an image to a target image while keep the content of the target image. Since NST is designed for general cases, it lacks the ability to focus on some special area for tasks such as cartoonization.
There are another family of methods based on Generative Adversarial Networks (GAN)~\cite{gan_first} that perform domain transfer in an adversarial manner. Some image2image translation methods (e.g., pix2pix~\cite{pix2pix}, Bicycle~\cite{zhu2017toward}) are proposed to map the image from one domain to another. However, these methods require paired images, which are difficult to obtain for many tasks. Thanks to a series of unsupervised domain transfer frameworks proposed (e.g., CycleGAN~\cite{cyclegan}, UNIT~\cite{unit}), we are able to train a model with unpaired data. There are already some existing methods (e.g., CartoonGAN~\cite{cartoongan}, DAGAN~\cite{dagan}) performing cartoonization based on unsupervised GAN framework, but they usually fail to capture the delicate facial parts or generate pleasing results.

According to our observation and analysis, we can find there exist three challenges in producing acceptable quality cartoon portraits. First, there are no public paired datasets on human selfies and cartoon portraits. Second, we do not know how to keep a cartoon style which includes abstract and simplified textural features. Last, it is difficult to generate delicate facial features when performing style transfer for portraits.

In this paper, we propose scGAN, a dedicated GAN designed for selfie cartoonization task. To address the above challenges, we first apply cycle consistency loss to solve unpaired training data. Then we proposed three simple yet effective loss functions, which can generate cartoon portraits from the overall outline to local details and keep the artistic form of the cartoon domain. The main contributions of this paper are as follows:
\begin{itemize}
\item We present scGAN, a novel selfie cartoonization method based on GAN, that can learn the mapping from real-world selfies to cartoon portraits in different styles.
\item We propose to utilize attention mechanism for cartoon portrait generation task based on the domain knowledge. It can be divided into two important aspects. On a pixel level, we introduce a total variation loss which forces the networks to only highlight important edge and content information. On a regional level, we propose an attentive cycle loss acting on target regions to generate more detailed facial features. In addition, we apply perceptual loss to improve object preservation ability and convergence properties during training.
\item We construct and release a new dataset for the cross-domain image generation task, which contains plenty of high-quality and different style cartoon portraits. It could play an important role in helping painters for selfie cartoonization by using a deep learning method.
\end{itemize}

\vspace{-0.3cm}
\section{RELATED WORK}
\subsection{Traditional Image Processing Methods}
Traditional methods are basically based on oversimplified mathematical theoretical models, such as~\cite{li2011guided,yang2010semantics}. These methods often apply face parsing methods to segment out each facial component, then use non-photorealistic rendering~\cite{kyprianidis2013state} method or simple filtering processing to obtain cartoon images. Based on these methods, there are various image cartoonization APPs on our mobile phone, such as MomanCamera, Cartoon Camera Photo Editor. Though achieving the goal of real-time processing in these APPs, they failed to generate detailed facial parts.
\vspace{-3mm}
\subsection{Deep Learning Methods}
NST has received considerable attention in artistic creation. But for our task, selfies usually contain complex and multiple facial structures. Generating cartoon portraits by this method would be more problematic due to the information separation of styles and contents. GAN is another popular method for domain transfer. Pix2pix~\cite{pix2pix} solves the image-to-image translation problems, but needs strict paired data. It is difficult to obtain such dataset for our selfie cartoonization task. To break this fundamental limitation, CycleGAN~\cite{cyclegan} enforces the one-to-one mapping with the help of cycle consistency loss. Based on this term, UNIT~\cite{unit} makes a shared-latent space assumption and proposes an unsupervised image translation framework. The latest study named CartoonGAN~\cite{cartoongan} performs well in generating high-quality cartoon images from real-world photos. However, all the above methods fail to generate satisfactory cartoon portraits for these two difficult points: (1) How to synthesize detailed facial features as far as possible. (2) How to keep the artistic characteristics (e.g.,smoothing areas and clear contents) of the cartoon images. As a comparison, our method outperforms these state-of-the-art methods.
\vspace{-0.3cm}
\section{PROPOSED METHOD}
Formally, the selfie cartoonization problem is formulated as learning a mapping function from domain $A$ formed by real selfies to domain $B$ formed by cartoon portraits. The mapping function is learned using training data $\{x_i\}_{i=1}^n$ in domain $A$ and $\{y_i\}_{i=1}^m$ in domain $B$, where $n$ and $m$ represent the numbers of two domain images. We denote the data distribution as $x \sim p_{\text{data}}(A)$ and $y \sim p_{\text{data}}(B)$. Like the classic GAN, we design the generative function $G$ that produces vivid images to confuse the discriminative function $D$, while the optimization procedure of $D$ aims to distinguish the real cartoon portraits in the domain $Y$ from the fake ones generated by $G$. Let $L$ be the loss function, $G_{*}$ and $D_{*}$ be the optimal weights of the networks, so we aim to solve the min-max problem:
\begin{equation}
(G_{*},D_{*})=\arg \min\limits_{G}\max\limits_{D}L(G,D).
\end{equation}
\vspace{-4mm}
\subsection{Network Architecture}
\begin{figure*}[!tp]
\begin{center}
\includegraphics[width=12cm]{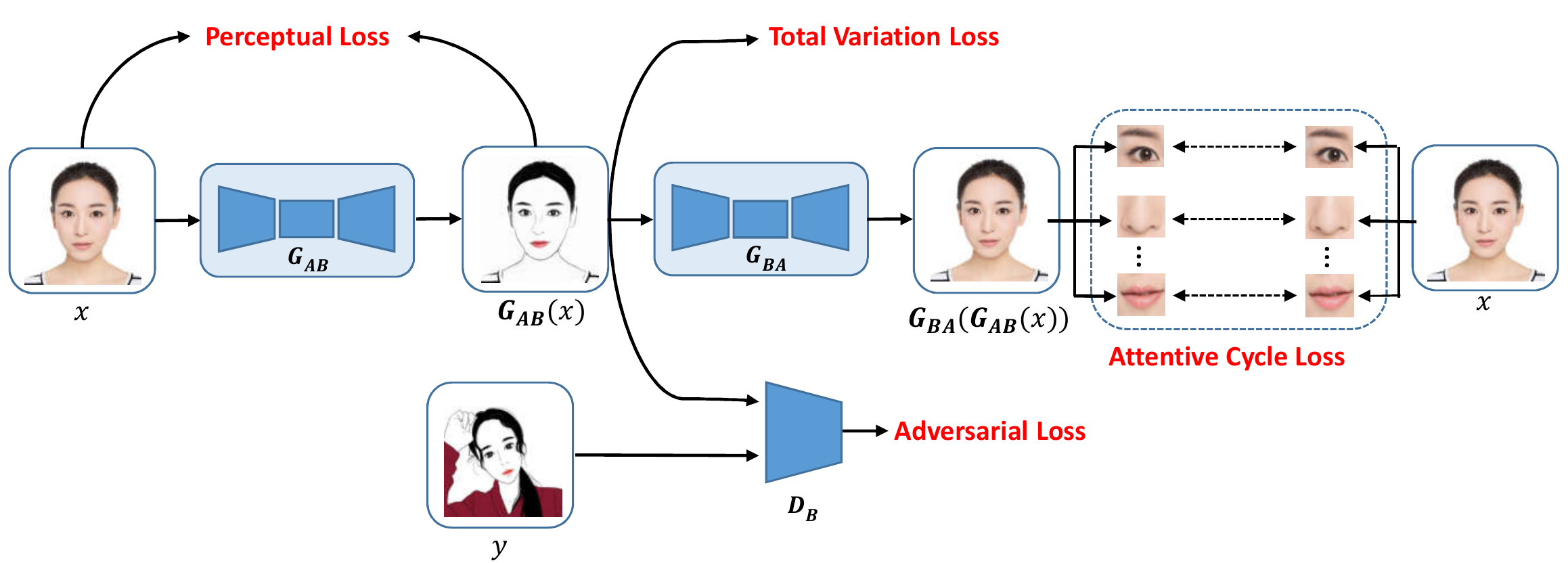}
\vspace{-4mm}
\caption{The framework of our scGAN. Besides the adversarial loss, we also propose three adaptations, namely total variation, attentive cycle and perceptual loss, based on the attention mechanism. For simplicity, we only present one direction transformation: selfie domain $\Rightarrow$ cartoon domain. We omit the other direction in this figure: cartoon domain $\Rightarrow$ selfie domain.}
\end{center}
\label{fig:fra}
\end{figure*}
Our network architecture is shown in Fig.~\ref{fig:fra}. Generator $G_{AB}$ firstly translates $x_i$ into a hand-painted cartoon image $G_{AB}(x_i)$, then generator $G_{BA}$ brings $G_{AB}(x_i)$ back to the original selfie. The other direction from selfie domain to cartoon domain is similar. For simplicity, We omit it in the Fig.~\ref{fig:fra}. The generator ($G_{AB}$) has an encoder-decoder structure. In consideration of the texture features of cartoon portraits and the size of datasets, we apply a Unet~\cite{unet} based architecture. As proved in pix2pix, Unet which is widely used in medical image segmentation can retain the prominent edges such as the eyelid. Moreover, to guarantee that the learned function is able to map an individual input $x_{i}$ to a designed output $y_{i}$, we reconstruct the input image with the same encoder-decoder structure ($G_{BA}$). For judging whether the generative result is a cartoon style image, we use a simple patch-level discriminator ($D_{B}$)~\cite{patchgan_first}. And we have a same discriminator ($D_{A}$) in the other direction.
\subsection{Adversarial Loss}
We apply the adversarial loss~\cite{gan_first} to both networks $G$ and $D$. For the mapping function $G_{AB}: A\rightarrow B$ and its discriminator $D_{B}$, we define the loss function as:
\begin{equation}
\begin{split}
L^{AB}_{GAN} &= \mathbb{E}_{y \sim P_{\text{data}} (B)}[\log D_B(y)] \\
&+ \mathbb{E}_{x \sim P_{\text{data}} (A)}[\log(1-D_B(G_{AB}(x)))],
\end{split}
\end{equation}
where the generative function $G_{AB}$ tries to produce hand-painted cartoon portraits which look the same as images from domain $B$, and meanwhile $D_{B}$ aims to distinguish between synthetic samples $G_{AB}(x)$ and real samples $y$. The mapping function $G_{BA}: B\rightarrow A$ and its discriminator $D_{A}$ are similar to the former.

\vspace{-0.3cm}
\subsection{Attentive Adversarial Network}
Besides the general adversarial loss, our scGAN applies attention mechanism for cartoonization task, which is described in the following parts.

\textbf{Region Level: Attentive Cycle Loss}. As discussed in CycleGAN, cycle consistency loss can break the reliance on paired data in domain transfer. On this basis, we propose an attentive cycle loss to guide the generator to lay more emphasis on detailed facial features, such as the slender eyelashes, attractive pupils and so on. In other words, doing so can capture the regions with great attention.

Specifically, given a real selfie $x_i$, we first apply a face parsing algorithm that can be a simple detector to detect each facial component, then obtain a set of attentive region location information, which is given by:
\begin{equation}
E(x)=[\text{bbox}_1,...,\text{bbox}_j,...,\text{bbox}_k],
\end{equation}
where $E(x)$ is a trained detector for detecting the facial components, $k$ denotes the number of attentive regions, and $bbox$ stands for the bounding boxes of each region. As a result, the more attentive parts we choose, the better results we get.
We use $x^{j}_{i}$ to express the region of image $x_i$ that is inside ${bbox}_{j}$. Therefore, the attentive cycle loss is defined as:
\begin{equation}
\begin{split}
L^{AB}_{att-cyc}&=\mathbb{E}_{{x\sim P}_{\text{data}}(A)}[\sum_{j=1}^k\lambda_{j}||G_{BA}(G_{AB}(x^{j}_{i}))-x^{j}_{i}||_{1}],
\end{split}
\end{equation}
$\lambda_{j}$ controls the relative importance of each facial component. Larger $\lambda$ drives the input images to maintain more regional information, and therefore, results in more detailed hand-painted cartoon portraits generation. In our experiments, we set $k = 4$, and $\lambda = [1, 0.5, 0.5, 0.5]$ represents the weights for [the whole image $x_i$, the eyes, the nose, the mouth] regions, respectively. Similarly, for the transformation from domain $B$ to $A$, we only adopt a standard cycle consistency loss as:
\begin{equation}
\begin{split}
L^{BA}_{cyc}&=\mathbb{E}_{{y\sim P}_{\text{data}}(B)}[||G_{AB}(G_{BA}(y))-y||_{1}].
\end{split}
\end{equation}

\begin{figure}[!ht]
\begin{center}
\includegraphics[scale=0.70]{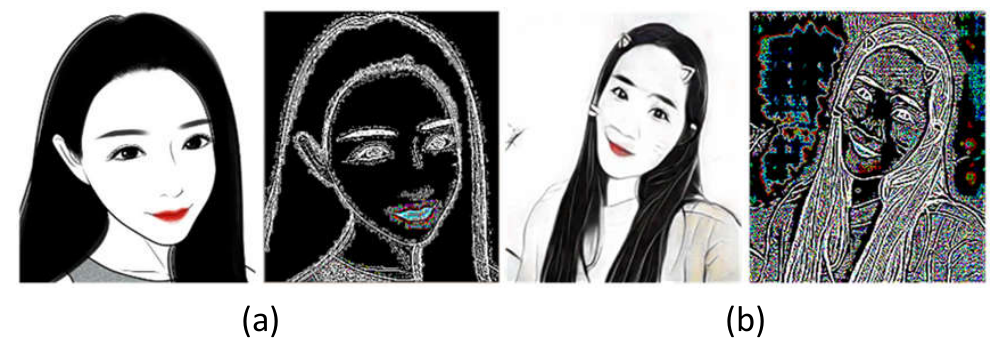}
\vspace{-5mm}
\caption{The gradient distribution maps. (a) is drawn by painter, (b) is generated without tv loss.}
\end{center}
\label{fig:tv}
\vspace{-5mm}
\end{figure}

\textbf{Pixel Level: Total Variation Loss}. Cartoon images usually have unique characteristics with high-level simplification and uniform color distribution. Based on this observation, we do one small experiment about image gradient distribution for hand-painted cartoon portraits. Fig.~\ref{fig:tv} provides a demonstration. Visually, there are fixed rules in gradient distribution maps for images drawn by painters. Gradient changes are not obvious in most regions except for the edge parts of facial structures. However, there are a complex set of unavoidable factors (e.g., illumination, wrinkle) in our selfies which would interfere transformation process severely.

In such case, one more important target is to keep edge parts and remove blurry factors. So we propose applying total variation as a loss function which only highlights important edge and content information in cartoon portraits. As discussed in~\cite{tv2015}, total variation as a regularizer method plays an important role in traditional image processing fields. But for obtaining faultless results we define total variation as a constraint function:
\begin{equation}
L_{tv} = \mathbb{E}_{x\sim P_{\text{data}} (A)}[|\nabla G_{AB}(x)|].
\end{equation}
In this way, we minimize the synthetic hand-painted portraits' gradient in the experiment, we find the results remove undesired details like the black regions of cheek as well as preserving important edge parts.

\textbf{Perceptual loss}. In many classical cases, a network with a low content loss (e.g., Mean Square Error) on each pixel leads to the blurry artifacts on generated results. In our method, rather than directly measuring differences between the input images and the output images using Euclidean distance, we apply perceptual loss to calculate differences between high-level feature maps extracted by 19-layer VGG networks pre-trained on ImageNet dataset. As has discussed in~\cite{johnson2016perceptual}, perceptual loss is more able to maintain image content and overall spatial structure when used as feature reconstruction loss. Compared with per-pixel differences, we find it has better object preservation ability, and it can accelerate convergence. Accordingly, we define this loss as:
\begin{equation}
L_{per} = \mathbb{E}_{x\sim P_{\text{data}} (A)}[||VGG_l(G_{AB}(x))- VGG_l(x)||_1],
\end{equation}
where $l$ represents the feature maps of a specific VGG layer. In our training, we choose the layer 'conv4\_4' to compute this loss.
\vspace{-0.3cm}
\subsection{Full Objective Function}
Overall, our full loss function is:
\begin{equation}
\begin{split}
L = L^{AB}_{GAN} + L^{BA}_{GAN}
&+ \alpha (L^{AB}_{att-cyc}+L^{BA}_{cyc})\\
+ \beta L_{tv}
+ \gamma L_{per},
\end{split}
\label{equ:all}
\end{equation}
where $\alpha,\beta,\gamma > 0$ are the weights to trade off different losses.
\section{EXPERIMENTS}
We implemented our scGAN in Pytorch. All experiments were performed on an NVIDIA P40 GPU. For our method, we set $\alpha$ = 10, $\beta$ = 2 and $\gamma$ = 0.5 in Equation~\ref{equ:all}.
\vspace{-0.3cm}
\subsection{Data collection}
The training data is made up of real selfies and cartoon portraits, and the test data only contains real selfies.
\textbf{Real selfies} are downloaded from Google Image Search using keywords (e.g., Women Portrait). We totally obtain 3,524 real selfies above the shoulder with the image cropping skills. 3,300 of these images for training and others for testing.
\textbf{Cartoon portraits} are composed of three different styles as shown in Table~\ref{tab:data}. These are difficult to obtain because there are no public datasets on human and cartoon portraits due to cost. So we can only download from online painting stores. However, different painters have different artistic styles and the most cartoon artworks contain digital image watermarks with concern about copyright. Taking these factors into consideration, we build the dataset by performing some preprocessing steps such as filtering, cropping etc.
\begin{table}[!ht]
\label{tab:data}
\begin{center}
\caption{The cartoon portraits dataset.}
\begin{tabular}{|c|c|} \hline
Style        & Numbers \\ \hline
Hand-painted &    850     \\
Watercolor   &    730    \\
Anime        &    3000   \\ \hline
\end{tabular}
\end{center}
\end{table}
\vspace{-8mm}
\subsection{Comparison with state-of-the-arts}
\begin{figure*}[t]
\begin{center}
\includegraphics[scale=0.75]{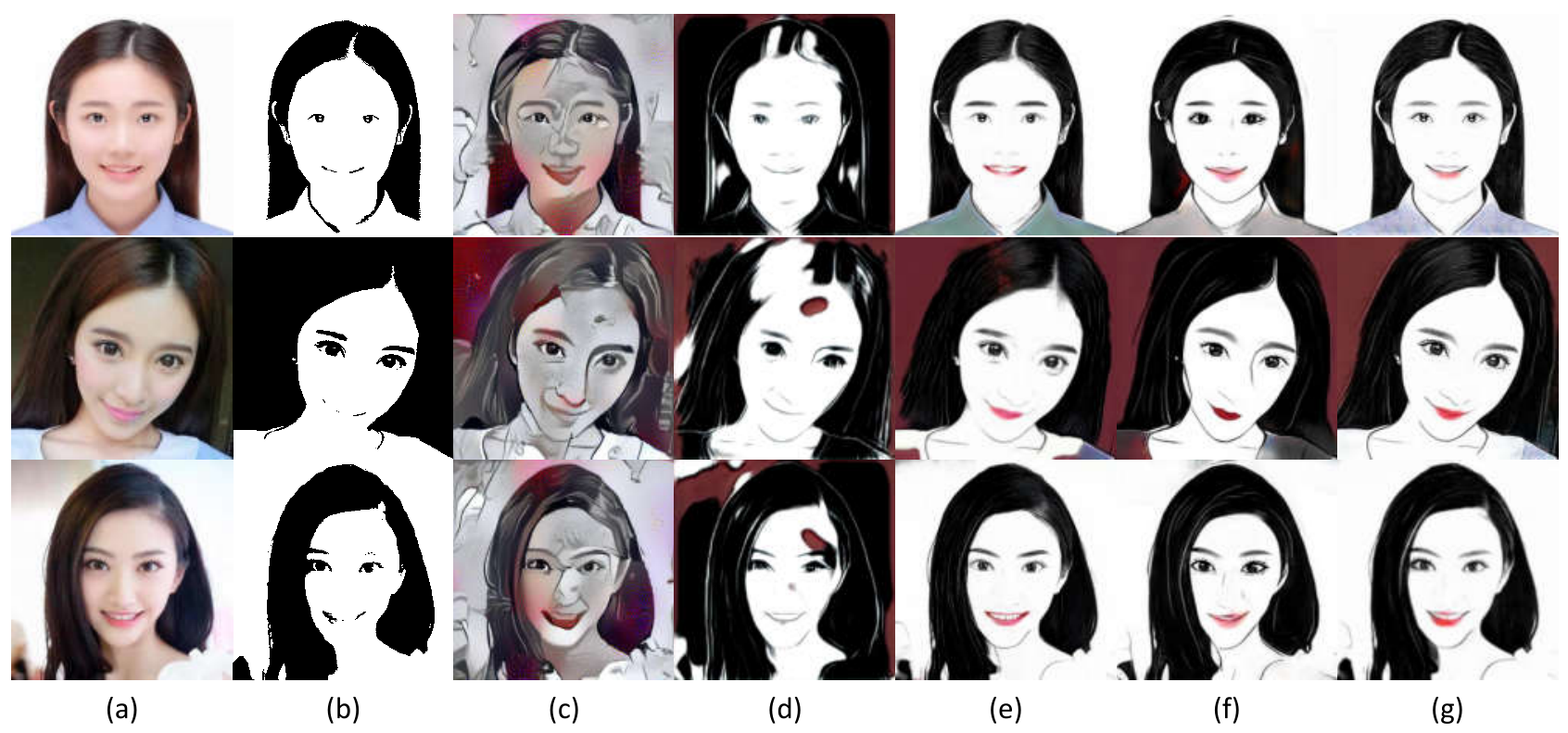}
\end{center}
\vspace{-4mm}
\caption{Results of comparing with different methods. (a) Input images, (b) Image binarization, (c) NST, (d) CartoonGAN, (e) UNIT, (f) CycleGAN, (g) Our results.}
\label{fig:all}
\end{figure*}

We present the comparative experiments between our method and representative stylization methods as follows:
\begin{itemize}
\item \textbf{Image Binarization} is a traditional image processing method which can obtain the black and white cartoon style portraits.
\vspace{-0.25cm}
\item \textbf{NST} \cite{nst} is a CNN-based stylization work. This method can transfer the artistic style of an image to a target image while keep the content of the target image.
\vspace{-0.25cm}
\item \textbf{UNIT} \cite{unit} is a recent unsupervised image2image translation method which based on the shared-latent space assumption.
\vspace{-0.25cm}
\item \textbf{CartoonGAN} \cite{cartoongan} is a latest work to transforming photos of the real-world scenes into cartoon style images.
\vspace{-0.6cm}
\item \textbf{CycleGAN} \cite{cyclegan} is a distinguished approach for learning to translate an image from a source domain $A$ to a target domain $B$ in the absence of paired examples.
\end{itemize}
Fig.~\ref{fig:all} shows the visual results. For lack of paired data, we run a perceptual study on this transformation. Specifically, we randomly shuffle all the methods and hide their name in Fig.~\ref{fig:all} then gather statistical results from 125 participants. According to the established grading rules (e.g., presentation style, overall quality, detail description and so on.), participants are asked to grade these generated cartoon images using a 5-point system, and a higher score represents more satisfaction overall. Table~\ref{tab:score} shows results regarding scoring system.

We can see that our proposed method succeeds in selfie cartoonization from overall structures to detailed facial features.
In more specific terms, image binarization as a basic method for image segmentation merely shows the rough facial shape.
As for NST, we select one image as $I_{s}$ randomly because all of the hand-painted cartoon portraits have little difference basically in style. We can see it fails to learn the style well, let alone keep delicate details of the local regions.
CartoonGAN as a dedicated GAN, the results are less than satisfactory in this task. We use the default parameters in literature for experiments. The stylization results of UNIT and CycleGAN perform slightly better which start to capture the major lineament at first glance, but the fine areas do have a large gap compared with our proposed method. In comparison, our method is able to generate correct colors, obtain sharper and detailed textures, and output overall more satisfactory hand-painted cartoon portrait results.

\begin{table}[!ht]
\caption{The results of our 5-point system. By taking the average, it shows that our method are most satisfactory.}
\vspace{-5mm}
\begin{center}
\setlength{\tabcolsep}{0.9 mm}
{
\begin{tabular}{|l|c|c|c|c|c|c|}
\hline
Methods\textbackslash{}Scores & 1 & 2 & 3 & 4 & 5 & $Avg.$ \\ \hline
\textbf{Binarization} & 20\% & 32.8\% & 32\% & 11.2\% & 4\% & 2.46 \\
\textbf{NST} & 56\%  & 20\% & 16.8\% & 4.8\% & 2.4\% & 1.78 \\
\textbf{CartoonGAN} & 48\% & 34.4\% & 13.6\%  & 1.6\% & 2.4\% & 1.76 \\
\textbf{UNIT}  & 3.2\% & 19.2\% & 35.2\%  & 38.4\% & 4\% & 3.21  \\
\textbf{CycleGAN} & 12\% & 24.8\%  & 34.4\% & 18.4\% & 10.4\% & 2.90  \\
\textbf{Our method} & 3.2\%& 12\% & 21.6\% & 33.6\% & 29.6\% & \textbf{3.74} \\ \hline
\end{tabular}}
\end{center}
\label{tab:score}
\end{table}

\vspace{-9mm}
\subsection{Ablation Study}
To verify the role of each part in our model, here we make ablation experiments to isolate the influence of each term.

\begin{figure}[!h]
\begin{center}
\includegraphics[scale=0.5]{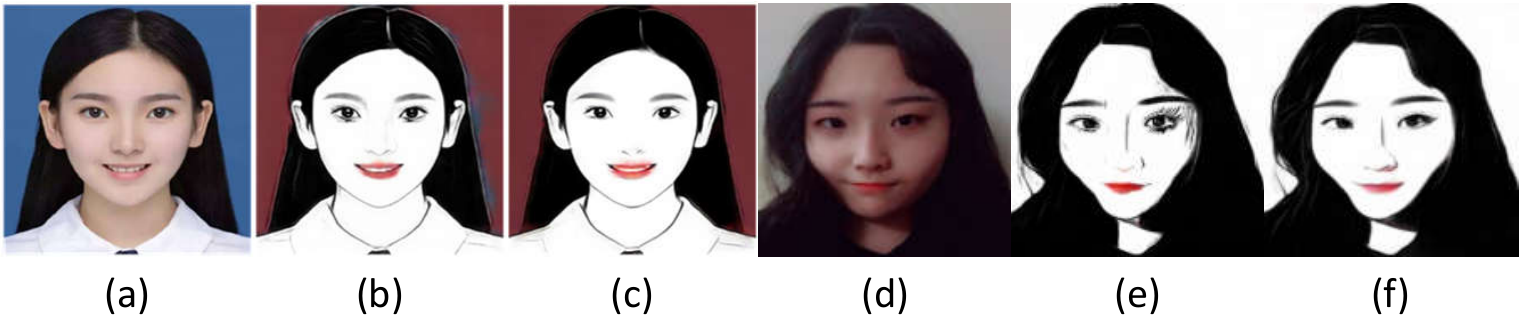}
\end{center}
\vspace{-5mm}
\caption{Results of the ablation study for tv loss. Input images: (a) and (d). Without tv loss: (b) and (e). With tv loss: (c) and (f).}
\label{fig:tv2}
\vspace{-5mm}
\end{figure}

\textbf{Total Variation Loss}. Fig.~\ref{fig:tv2} illustrates some examples of how tv loss influences the result. For input images (a) and (d), the results without tv loss are shown in (b) and (e), which show noisy and inconsistent regions. After using tv loss, we obtain results like (c) and (f), which are much smoother and clean. The intuition behind is that tv loss, as a regularizer, would force the network to highlight important edge information (e.g., delicate eyeliner) and remove dirty factors (e.g., clean cheek). To further investigate, we also analyze the change of the images' average gradient, as shown in Table~\ref{tab:gra}, and we observe that minimizing the gradients always leads to visually better results. For real life applications, there is often inevitable illumination or complex background influence, therefore, it is very necessary to impose constraints such as tv loss to obtain more satisfactory results.
\vspace{-5mm}
\begin{table}[!h]
\caption{The average gradient performances for generated results in Fig.~\ref{fig:tv2}.}
\begin{center}
\begin{tabular}{c|cc}
\hline
index & w/o tv loss  & w/ tv loss \\ \hline
(1)         & 36.5       & 30.9          \\
(2)         & 42.8       & 36.8          \\ \hline
\end{tabular}
\end{center}
\label{tab:gra}
\vspace{-5mm}
\end{table}

\begin{figure}[!ht]
\begin{center}
\includegraphics[scale=0.45]{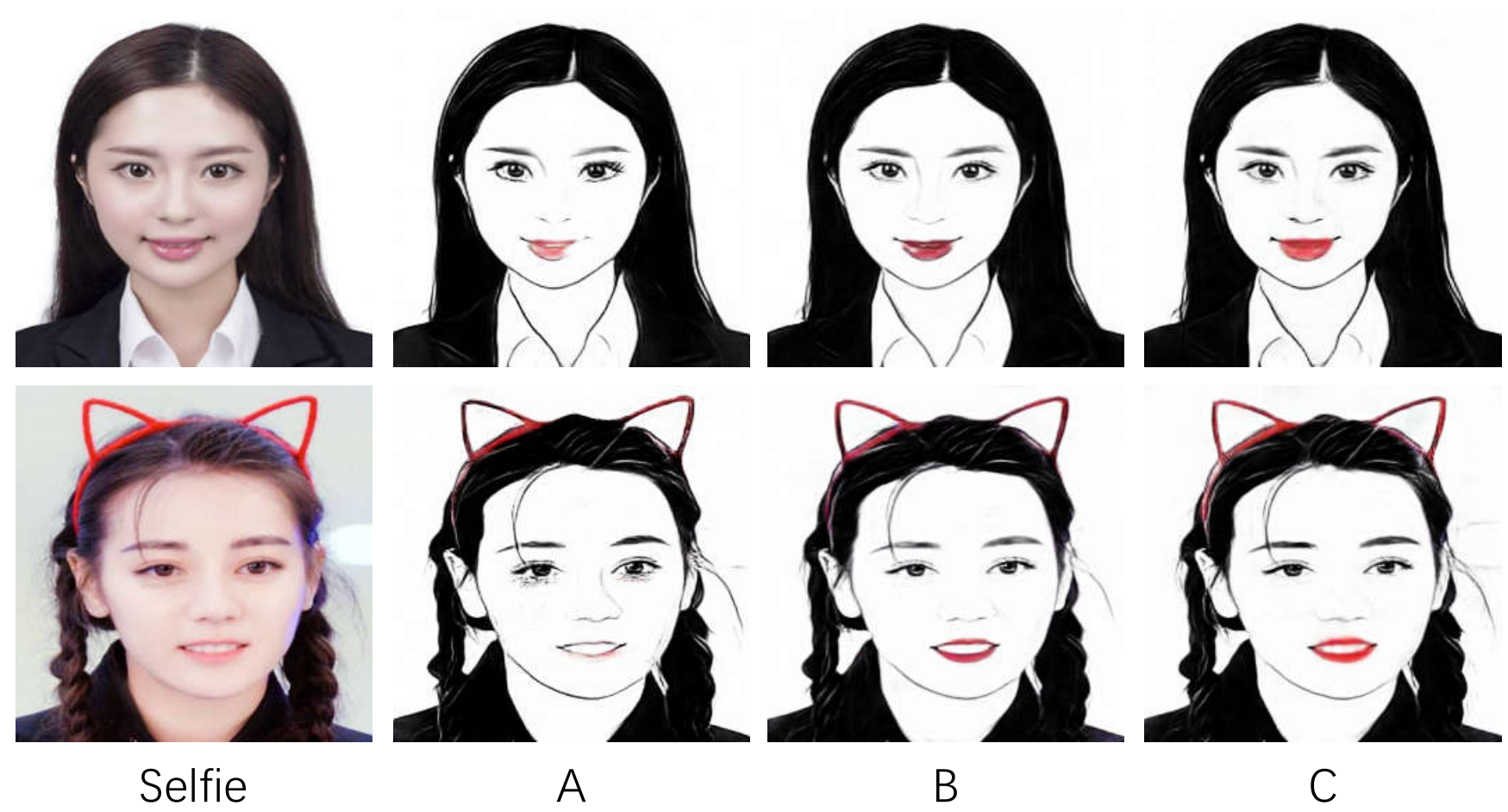}
\end{center}
\vspace{-5mm}
\caption{Results of the ablation study. The first column are input selfies. Each row in the remaining columns shows the cartoonization results of three experiments: A, B, C.}
\label{fig:abl}
\end{figure}

\textbf{Attentive cycle and Perceptual loss}. We also conduct experiments to study the impact of attentive cycle and perceptual loss. The experiment setup is described in Table~\ref{table:set} and some example outputs are illustrated in Fig.~\ref{fig:abl} correspondingly. Using only the CycleGAN loss (Experiment A) does not give satisfactory results, as shown in second column of Fig.~\ref{fig:abl}. The model trained with our attentive cycle loss (Experiment B) shows more potential high quality facial parts, eyes, nose and mouth. In Experiment C, adding the perceptual loss further improves the results by better preserving low-level features (e.g., the overall eyebrows and the color in mouth).
\begin{table}[ht]
\caption{Ablation study setups.}
\begin{center}
\begin{tabular}{|cccc|}
\hline
Setup & $L_{cyc}$ & $L\_{att\_cyc}$ & $L_{per}$ \\ \hline
A     &     \checkmark     &      &      \\
B     &         &   \checkmark   &      \\
C     &         &   \checkmark   & \checkmark     \\ \hline
\end{tabular}
\end{center}
\label{table:set}
\end{table}

\subsection{Anime and Watercolor Style Generation}
\begin{figure}[!ht]
\begin{center}
\includegraphics[width=80mm]{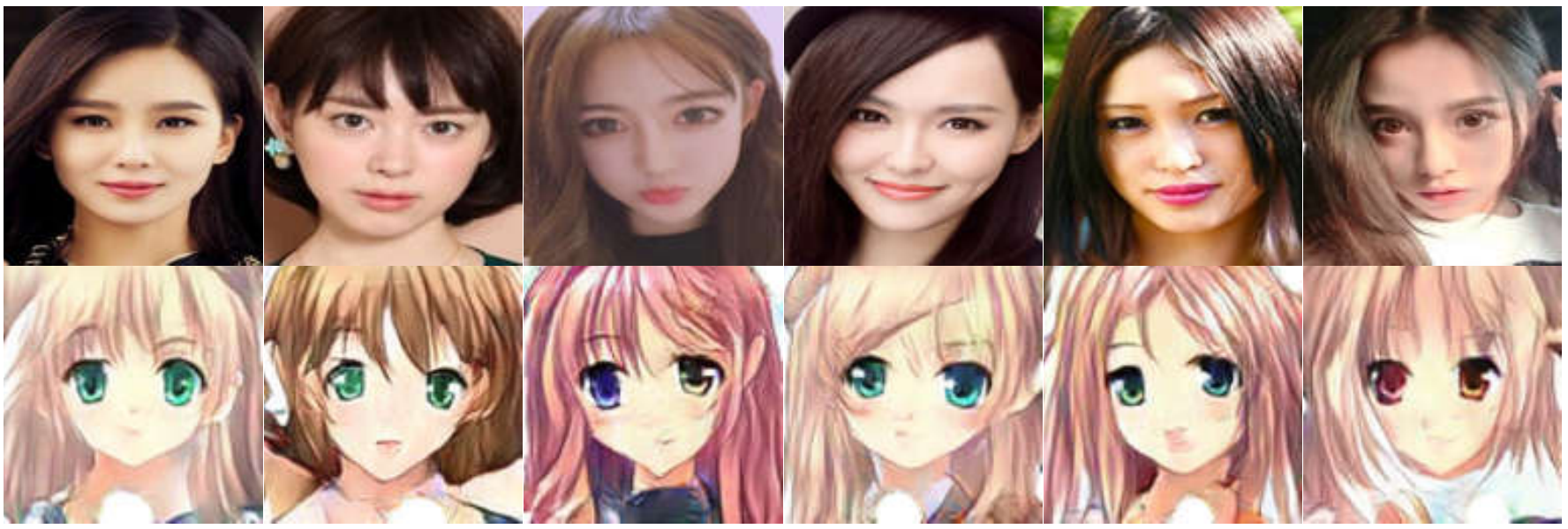}
\end{center}
\vspace{-5mm}
\caption{Results of selfie cartoonization for anime style.}
\label{fig:anime}
\end{figure}

\begin{figure}[!h]
\begin{center}
\includegraphics[width=80mm]{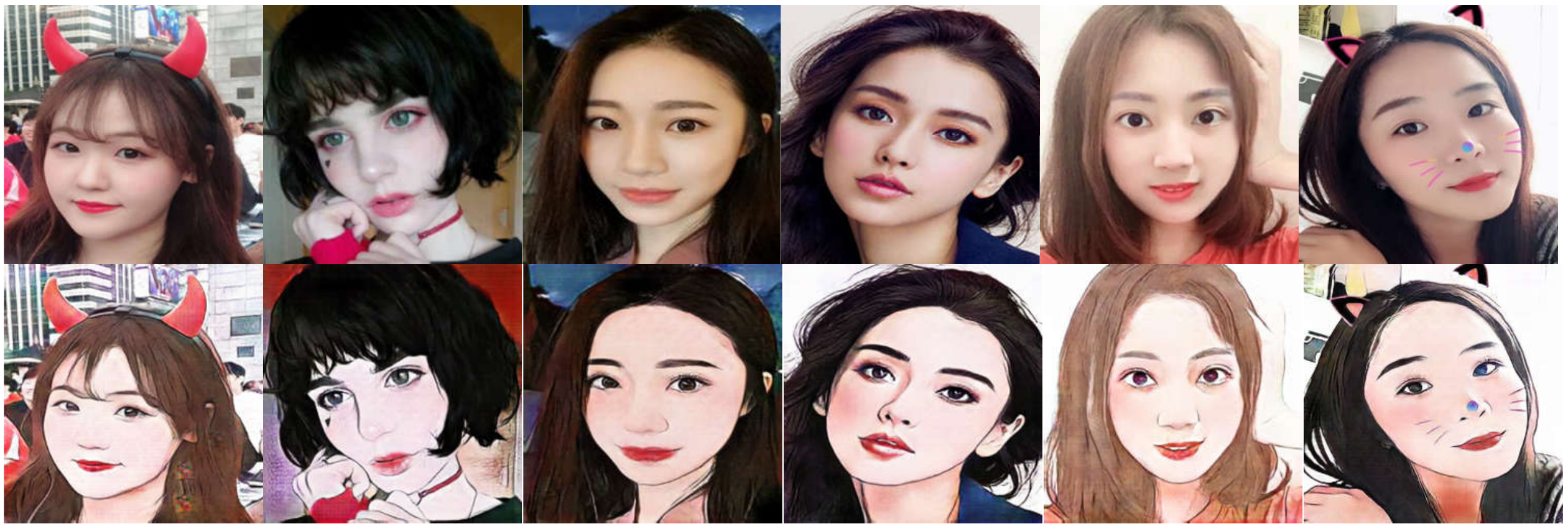}
\end{center}
\vspace{-5mm}
\caption{Results of selfie cartoonization for watercolor style.}
\label{fig:watercolor}
\end{figure}

We also attempt to translate real selfies into other styles, the results are shown in Fig.~\ref{fig:anime}. and Fig.~\ref{fig:watercolor}.

\section{Conclusion}
In this paper, we propose scGAN, a novel method for selfie cartoonization with attentive adversarial networks. Based on domain knowledge, we propose to utilize attention mechanism for cartoon domain generation. In addition, we construct and release a new dataset for domain transformation task between selfies and cartoon portraits.
In the future work, we would like to investigate how to generate full body cartoon images with high quality and efficiency. Furthermore, how to handle videos in real time for supporting the interesting applications is another promising research direction.

\textbf{Acknowledgement}. This work was partially funded by National Natural Science Foundation of China NO.61602463, and the Open Project Program of the National Laboratory of Pattern Recognition (NLPR).

\bibliographystyle{IEEEbib}
\bibliography{icme2019template}

\end{document}